# Performance evaluation of wavelet scattering network in image texture classification in various color spaces

Jiasong Wu[1,4]  Longyu Jiang[1,4]  Xu Han[1,4]  Lotfi Senhadji[2,3,4]  Huazhong Shu[1,4]

([1]Laboratory of Image Science and Technology, the Key Laboratory of Computer Network and Information Integration, Southeast University, Ministry of Education, Nanjing, China)
([2]INSERM, U 1099, Rennes 35000, France)
([3]Laboratoire Traitement du Signal et de l'Image, Université de Rennes 1, Rennes 35000, France)
([4]Centre de Recherche en Information Biomédicale Sino-français, Nanjing 210096, China)
*E-mail: jswu@seu.edu.cn; jly@seu.edu.cn; xuhan@seu.edu.cn; lotfi.senhadji@univ-rennes1.fr; shu.list@seu.edu.cn*

**Abstract:** Texture plays an important role in many image analysis applications. In this paper, we give a performance evaluation of color texture classification by performing wavelet scattering network in various color spaces. Experimental results on the KTH_TIPS_COL database show that opponent RGB based wavelet scattering network outperforms other color spaces. Therefore, when dealing with the problem of color texture classification, opponent RGB based wavelet scattering network is recommended.

**T**exture plays an important role in many image analysis techniques such as image segmentation, image retrieval, and image classification. Due to its importance, texture feature description has attracted much attention in the past decades. Lowe proposed a very famous data-dependent local feature descriptor, namely, scale-invariant feature transform (SIFT) [1, 2], which computes the local sum of image gradient amplitudes among image gradients having nearly the same direction in a histogram with eight different direction bins. Tola *et al*. [3] proposed a data-independent descriptor, known as DAISY, to approximate SIFT coefficients by $S[\lambda_1]x = |x * \psi_{\lambda_1}| * \phi_{2^J}(u)$, where $\psi_{\lambda_1}$ are partial derivatives of a Gaussian computed at the finest image scale, along eight different rotations. The averaging filter $\phi_{2^J}$ is a scaled Gaussian. Mallat and Bruna [4, 5] proposed a new data-independent feature descriptor, that is, wavelet scattering networks (ScatNet), which can be seen as a multilayer version of DAISY, substituting the partial derivatives of a Gaussian by complex wavelets. Note that the above descriptors are mainly proposed for grey-scale images.

As we know, color provides useful information for image classification and object recognition [6-9]. For example, Gevers and Smeulders [6] proposed color invariants to a substantial change in viewpoint, object geometry and illumination. Drimbarean and Whelan [7] extended three grey scale descriptors (local linear transforms, Gabor filtering, and co-occurrence) to color ones and determined the contribution of color information to the texture classification performance. Maenpaa and Pietikainen [8] performed four descriptors (color histograms, color ratio histograms, Gabor filtering, local binary pattern) in various color spaces for color texture classification. Van de Sande *et al*. [9] evaluated various color descriptors, including color histogram, color moments, and various color SIFT descriptors, for object and scene recognition. They concluded that opponent SIFT is recommended when choosing a single descriptor and no prior knowledge about the data set and object and scene categories is available. Zhang *et al*. [10] further extended the idea of opponent SIFT to double opponent SIFT, which took Red-Cyan (R-C) channel into consideration. Their method provides better results. Oyallon *et al*. [11] extended the grey-scale ScatNet descriptor to color one, however, they only considered the YUV color space.

In this paper, we will consider wavelet scattering networks in various color spaces, and try to find the best color spaces for the utilization of wavelet scattering networks.

## 1. Color Spaces

Table 1 lists the color spaces that will be considered in the paper. Notice that we only consider the color spaces that can be easily obtained from RGB. According to Table 1, some typical colors spaces are chosen from each category. Equations (1)-(6) show the color spaces conversions from RGB to YCbCr, HSL, $I_1I_2I_3$, CIE XYZ, opponent RGB and double opponent RGB, respectively.

**Table 1 Various color spaces and their brief descriptions (six categories)**

| Color spaces | Descriptions |
|---|---|
| RGB | Red, Green, Blue |
| YUV | |
| YIQ | |
| YPbPr | Luminance, Chrominance |
| YCbCr | |
| JPEG-YCbCr | |
| YDbDr | |
| HSV | Hue, Saturation, Value |
| HSL | Hue, Saturation, Lightness |
| HSI | Hue, Saturation, Intensity |
| $I_1I_2I_3$ | Linear transform of RGB |
| CIE XYZ | |
| CIE LUV | |
| CIE LCH | Tristimuli, Chromaticity, |
| CIE LAB | and Colorimetric systems |
| CIE UVW | |
| CAT02 LMS | |
| Opponent RGB[9] | Opponent theory |
| Double Opponent RGB[10] | |

1) RGB to YCbCr:
$$\begin{bmatrix} Y \\ C_b \\ C_r \end{bmatrix} = \frac{1}{256}\begin{bmatrix} 65.481 & 128.553 & 24.966 \\ -37.797 & -74.203 & 112.0 \\ 112.0 & -93.786 & -18.214 \end{bmatrix}\begin{bmatrix} R \\ G \\ B \end{bmatrix} + \begin{bmatrix} 16 \\ 128 \\ 128 \end{bmatrix}, \quad (1)$$

2) RGB to HSL:
$$r = \frac{R}{R+G+B}, g = \frac{G}{R+G+B}, b = \frac{B}{R+G+B}$$
$$\max = \max(r,g,b), \quad \min = \min(r,g,b)$$

$$H = \begin{cases} 0°, \text{if } \max = \min \\ 60° \times \frac{g-b}{\max-\min} + 0°, \text{if } \max = r \text{ and } g \geq b \\ 60° \times \frac{g-b}{\max-\min} + 360°, \text{if } \max = r \text{ and } g < b \\ 60° \times \frac{b-r}{\max-\min} + 120°, \text{if } \max = g \\ 60° \times \frac{r-g}{\max-\min} + 240°, \text{if } \max = b \end{cases}$$

$$L = \frac{1}{2}(\max+\min),$$

$$S = \begin{cases} 0, \text{ if } L = 0 \text{ or } \max = \min \\ \frac{\max-\min}{\max+\min} = \frac{\max-\min}{2L}, \text{if } 0 < L \leq \frac{1}{2} \\ \frac{\max-\min}{2-(\max+\min)} = \frac{\max-\min}{2-2L}, \text{if } L > \frac{1}{2} \end{cases}$$
(2)

3) RGB to $I_1I_2I_3$:
$$\begin{bmatrix} I_1 \\ I_2 \\ I_3 \end{bmatrix} = \begin{bmatrix} 1/3 & 1/3 & 1/3 \\ 1/2 & 0 & -1/2 \\ -1/4 & 1/2 & -1/4 \end{bmatrix}\begin{bmatrix} R \\ G \\ B \end{bmatrix}, \quad (3)$$

4) RGB to CIE XYZ:
$$\begin{bmatrix} X \\ Y \\ Z \end{bmatrix} = \frac{1}{0.177}\begin{bmatrix} 0.49 & 0.31 & 0.20 \\ 0.177 & 0.812 & 0.011 \\ 0.00 & 0.01 & 0.99 \end{bmatrix}\begin{bmatrix} R \\ G \\ B \end{bmatrix}, \quad (4)$$

5) RGB to Opponent RGB [9]:
$$\begin{bmatrix} O_{11} \\ O_{12} \\ O_{13} \end{bmatrix} = \begin{bmatrix} 1/\sqrt{2} & -1/\sqrt{2} & 0 \\ 1/\sqrt{6} & 1/\sqrt{6} & -2/\sqrt{6} \\ 1/\sqrt{3} & 1/\sqrt{3} & 1/\sqrt{3} \end{bmatrix}\begin{bmatrix} R \\ G \\ B \end{bmatrix}, \quad (5)$$

6) RGB to double Opponent RGB [10]:
$$\begin{bmatrix} O_{21} \\ O_{22} \\ O_{23} \\ O_{24} \end{bmatrix} = \begin{bmatrix} 1/\sqrt{2} & -1/\sqrt{2} & 0 \\ 2/\sqrt{6} & -1/\sqrt{6} & -1/\sqrt{6} \\ 1/\sqrt{6} & 1/\sqrt{6} & -2/\sqrt{6} \\ 1/\sqrt{3} & 1/\sqrt{3} & 1/\sqrt{3} \end{bmatrix}\begin{bmatrix} R \\ G \\ B \end{bmatrix}. \quad (6)$$

For more information about color spaces, we refer to [12, 13].

## 2. Method

Let the complex band-pass filter $\psi_\lambda$ be constructed by scaling and rotating a filter $\psi$ respectively by $2^j$ and $\theta$, that is,
$$\psi_\lambda(x) = 2^{2j}\psi(2^j\theta^{-1}x), \quad \lambda = 2^j\theta, \quad (7)$$
with $0 \leq j \leq J-1$, and $\theta = k\pi/K, \ k = 0,1,...,K-1$.
The wavelet-modulus coefficients of *f(x)* are given by
$$U[\lambda]f(x) = |f * \psi_\lambda(x)|. \quad (8)$$
The scattering propagator U[p] is defined by cascading wavelet-modulus operators [4]

$$U[p]f(x) = U[\lambda_m]\cdots U[\lambda_2]U[\lambda_1]f(x)$$
$$= \left\| |f * \psi_{\lambda_1}| * \psi_{\lambda_2} |\cdots * \psi_{\lambda_m}(x)\right\|, \quad (9)$$

where $p = (\lambda_1, \lambda_2, ..., \lambda_m)$ are the frequency-decreasing paths, that is, $|\lambda_k| \geq |\lambda_{k+1}|, k = 1, 2, ..., m-1$.

Scattering operator $S_J$ performs a spatial averaging on a domain whose width is proportional to $2^J$:

$$S_J[p]f(x) = U[p]f * \phi_J(x)$$
$$= U[\lambda_m]\cdots U[\lambda_2]U[\lambda_1]f * \phi_J(x) \quad (10)$$
$$= \left\| |f * \psi_{\lambda_1}| * \psi_{\lambda_2} |\cdots * \psi_{\lambda_m}(x)\right\| * \phi_J(x).$$

The wavelet scattering network is shown in Fig. 1. The network nodes of the layer $m$ correspond to the set $P^m$ of all paths $p = (\lambda_1, \lambda_2, ..., \lambda_m)$ of length $m$. This $m$th layer stores the propagated signals $\{U[p]f(x)\}_{p \in P^m}$ and outputs the scattering coefficients $\{S_J[p]f(x)\}_{p \in P^m}$. The output is obtained by cascading the scattering coefficients of every layers.

We apply wavelet scattering network to each channel of color images, and then, cascading the scattering coefficients of each channel to form the final outputs.

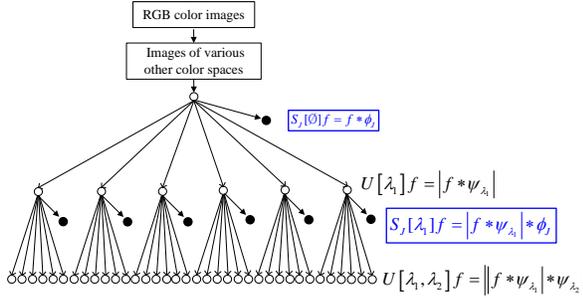

**Fig. 1 Three-layer wavelet scattering network. "○" denotes applying scattering propagator U to obtain the intermediate results. "●" denotes the outputs of the scattering network (scattering coefficients).**

## 3. Numerical experiment

In this Section, we perform wavelet scattering network in various color spaces on the KTH_TIPS_COL database [14], which contains 10 classes of 81 color images with controlled scaling, shear and illumination variations, but without rotation. Some examples are shown in Fig. 2.

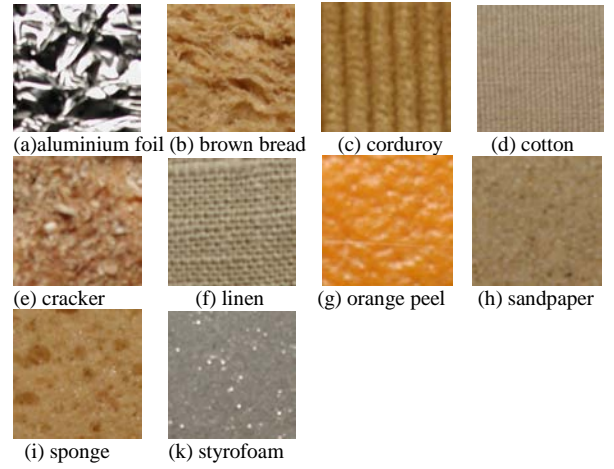

(a) aluminium foil (b) brown bread (c) corduroy (d) cotton
(e) cracker (f) linen (g) orange peel (h) sandpaper
(i) sponge (k) styrofoam

**Fig. 2 Some color image examples of KTH_TIPS_COL database. Color image size is 200×200×3.**

Numerical experiment is performed as follows:
1) Transfer all the color images on the KTH_TIPS_COL database from RGB space to various other color spaces that are shown in Table 1.
2) Randomly divide the total number of images of KTH_TIPS_COL database into two parts, that is, training data and testing data. In our experiment, we randomly choose 41 images for training and the rest 40 images for testing in each class. Therefore, we have totally 410 training images and 400 testing images.
3) Initialize the parameters of wavelet scattering network. We construct three-layer scattering network by complex Gabor wavelets, whose finest scale is $2^J = 16$ and total number of angles is K = 8. Oversampling factor is set to $2^1 = 2$. The corresponding scaling function covering the low frequency bands is a Gaussian.
4) Put every images of training set into wavelet scattering network, which is shown in Fig. 1, and obtain the training characteristic matrix $Q_0$, each column of which stores a scattering vector corresponding to one image. The number of columns of the matrix corresponds to the number of training images. Each channels of color image is computed separately and their scattering coefficients are concatenated. For each channel of color image with size 200×200, we have a scattering vector of 417 dimensions. Therefore, for 410 training color images with three channels, we obtain a training characteristic matrix $Q_0$ of size 1251×410.

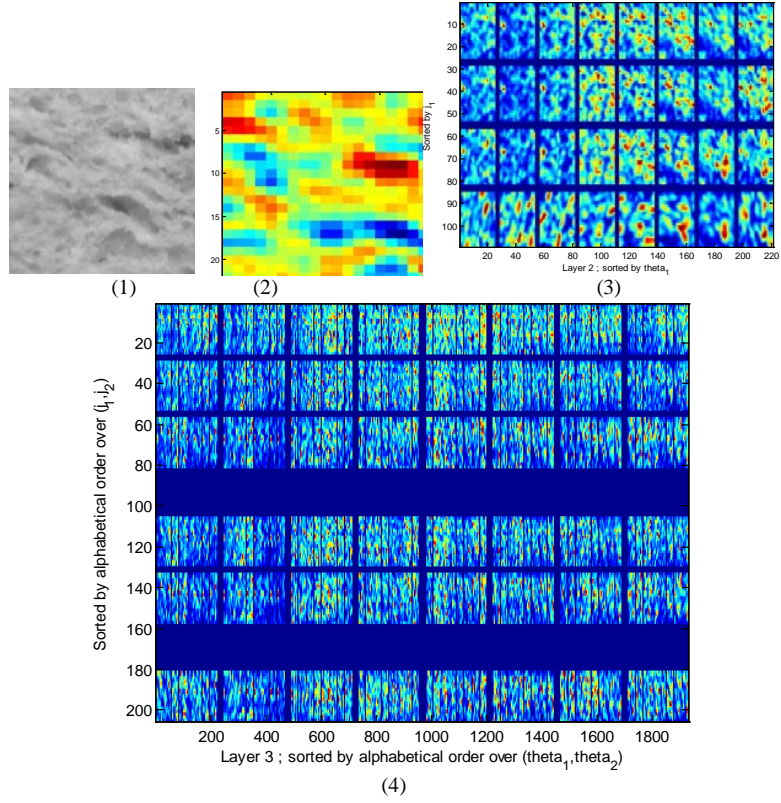

(1)　(2)　(3)

(4)

**Fig. 3 Scattering coefficients of R-channel of brown_bread**

**Table 2 Classification accuracies (%) of wavelet scattering network in various color spaces on the KTH_TIPS_COL [14] database. Best results are bolded. "Dim", "Accu", and "Col" denote the dimensions of principal components, classification accuracies, and color spaces, respectively.**

| Accu \ Dim \ Col | 5 | 10 | 15 | 20 | 25 | 30 | 35 | 40 | 45 | 50 |
|---|---|---|---|---|---|---|---|---|---|---|
| RGB | 92.98 | 97.58 | 97.85 | 98.55 | 99.08 | 98.73 | 98.93 | 98.17 | 98.65 | 98.55 |
| YUV | 91.95 | 97.3 | 98.05 | 98.55 | 98.78 | 98.65 | 98.63 | 98.48 | 98.63 | 98.17 |
| YIQ | 93.05 | 97.65 | 97.65 | 98.53 | 98.38 | 98.55 | 98.68 | 98.33 | 98.68 | 98.38 |
| YPbPr | 92.87 | 97.45 | 97.88 | 98.30 | 98.68 | 98.43 | 98.40 | 98.75 | 98.55 | 98.53 |
| YCbCr | 92.48 | 97.47 | 98.08 | 98.48 | 98.60 | 98.50 | 98.53 | 98.73 | 98.53 | 98.88 |
| JPEG-YCbCr | 92.28 | 96.80 | 97.75 | 98.45 | 98.43 | 98.60 | 98.55 | 98.22 | 98.95 | 98.45 |
| YDbDr | 91.62 | 97.23 | 98.15 | 98.53 | 98.28 | 98.28 | 98.48 | 98.75 | 98.60 | 98.78 |
| HSV | 85.42 | 91.55 | 95.40 | 95.50 | 96.10 | 96.78 | 96.00 | 96.15 | 95.25 | 96.22 |
| HSL | 84.62 | 91.85 | 94.15 | 95.82 | 96.43 | 96.60 | 96.55 | 96.15 | 96.15 | 96.12 |
| HSI | 83.68 | 90.97 | 96.12 | 95.82 | 96.55 | 96.50 | 96.00 | 96.32 | 95.70 | 96.05 |
| $I_1I_2I_3$ | 92.98 | 97.25 | 97.98 | 98.45 | 98.30 | 98.75 | 98.48 | 98.28 | 98.43 | 98.13 |
| CIE XYZ | 93.83 | 97.60 | 98.00 | 98.25 | 98.15 | 98.55 | 98.73 | 98.68 | 98.65 | 98.65 |
| CIE LUV | 92.43 | 96.80 | 97.68 | 98.20 | 98.73 | 98.78 | 98.43 | 98.93 | 98.60 | 98.53 |
| CIE LCH | 92.70 | 97.05 | 97.70 | 98.45 | 98.53 | 98.23 | 98.43 | 98.93 | 98.70 | 98.75 |
| CIE LAB | 92.88 | 97.45 | 98.33 | 98.65 | 98.43 | 98.58 | 98.73 | 98.50 | 99.00 | 98.48 |
| CAT02 LMS | 93.50 | 97.20 | 97.82 | 98.53 | 98.83 | 98.78 | 98.88 | 98.35 | 98.48 | 98.80 |
| Opponent RGB | **94.55** | **98.93** | 98.95 | 99.28 | **99.53** | 99.33 | **99.38** | 99.00 | 99.43 | 99.38 |
| Double Opponent RGB | 94.00 | 98.70 | **99.18** | **99.33** | 99.40 | **99.60** | 99.35 | **99.35** | **99.45** | **99.45** |

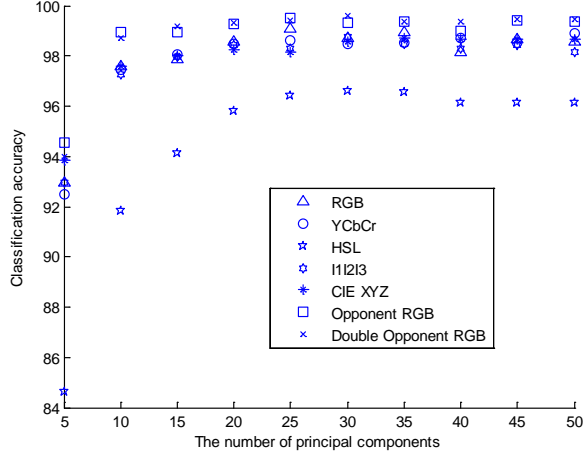

**Fig. 4 Classification accuracy of wavelet scattering network in some typical color spaces on the KTH_TIPS_COL [12] database.**

Let us take the image "brown bread" (Fig. 2 (b)) as example. Fig. 3(a) shows the R-channel image of brown bread. Fig. 3(b)-(d) shows the first, second, and third layer scattering coefficients, respectively. Note that we use the Matlab package "scatnet-0.2" which is available on the website [15].

5) Put every images of testing set into wavelet scattering network (Fig. 1), and obtain the testing characteristic matrix $Q_1$. Similar to step 4, for 400 testing images, we get the testing characteristic matrix $Q_1$ of size 1251×400.

6) Concatenate matrices $Q_0$ and $Q_1$, and obtain the scattering characteristic matrix Q=[$Q_0$, $Q_1$], whose size is 2151×810, which represents the characteristics of KTH_TIPS_COL database.

7) Put matrix **Q** into linear principal component analysis (PCA) classifier, choose the dimensions of principal component, and obtain the classification results, which are shown in Table 2. The classification results are averaged over 10 different random splits.

To show the results more clearly, we plot the classification results for some typical color spaces (RGB, YCbCr, HSL, $I_1I_2I_3$, CIE XYZ, Opponent RGB, Double Opponent RGB) in Fig. 4. From Table 2 and Fig. 4, we can see that opponent based color descriptors (Opponent RGB, Double Opponent RGB) outperform other color spaces. Luminance-Chrominance color spaces (YUV, YIQ, YPbPr, YCbCr, JPEG-YCbCr, YDbDr), CIE color systems (XYZ, LUV, LCH, LAB, CAT02 LMS) and $I_1I_2I_3$ provide similar classification accuracies. Hue-Saturation color spaces (HSV, HSL, HSI) achieve the worst classification accuracy. From the results, we also observe that the opponent RGB based color space is more suitable for color texture classification than YUV color space as reported in [11].

For observing some useful information clearly, we also plot the classification results in some typical color spaces (RGB, YCbCr, HSL, $I_1I_2I_3$, CIE XYZ, Opponent RGB, Double Opponent RGB) in Fig. 4. From Table 2 and Fig. 4, we can see that opponent based color descriptors (Opponent RGB, Double Opponent RGB) are superior to that of other color spaces. Luminance-Chrominance color spaces (YUV, YIQ, YPbPr, YCbCr, JPEG-YCbCr, YDbDr), CIE color systems (XYZ, LUV, LCH, LAB, CAT02 LMS) and $I_1I_2I_3$ achieve similar classification accuracies. Hue-Saturation color spaces (HSV, HSL, HSI) achieve the worst classification accuracy. It seems that the color spaces with linear transforms of RGB is better than that of nonlinear transforms of RGB when dealing with color image texture classification.

## 4. Conclusions

In this paper, we give a comparison study of wavelet scattering networks in various color spaces for color texture classification on the KTH_TIPS_COL database. Opponent based color descriptors are superior to that of other color spaces. Therefore, when choosing a single descriptor and no prior knowledge about the data set is available, the opponent RGB (or double opponent RGB) wavelet scattering network is recommended and also Hue-Saturation based color spaces (HSV, HSL, HSI) are not recommended.

**Acknowledgements**
This work was supported by the National Basic Research Program of China under Grant 2011CB707904, by the NSFC under Grants 61201344, 61271312, 11301074, and by the SRFDP under Grants 20110092110023 and 20120092120036, the Project-sponsored by SRF for ROCS, SEM, and by Natural Science Foundation of Jiangsu Province under Grant BK2012329 and by Qing Lan Project. This work is supported by INSERM postdoctoral fellowship.